\documentclass[11pt]{article}
\usepackage{coling2020}
\usepackage{times}
\usepackage{url}
\usepackage{latexsym}
\usepackage{enumitem}
\setlist[itemize]{noitemsep, nolistsep}
\usepackage{float}
\usepackage{graphicx}

\colingfinalcopy 

\title{JUNLP at SemEval-2020 Task 9: Sentiment Analysis of Hindi-English code mixed data using Grid Search Cross Validation}
\newcommand\Mark[1]{\textsuperscript#1}
\author{Avishek Garain\Mark{1},
Sainik Kumar Mahata\Mark{2}, Dipankar Das\Mark{3}\\
\Mark{1}\Mark{2}\Mark{3}Jadavpur University, Kolkata\\
\Mark{1}avishekgarain@gmail.com,
\Mark{2}sainik.mahata@gmail.com,\\
\Mark{3}dipankar.dipnil2005@gmail.com}

\date{}

\begin{document}
\maketitle
\begin{abstract}
\blfootnote{\noindent This work is licensed under a Creative Commons Attribution 4.0 International License. License details:\\
http://creativecommons.org/licenses/by/4.0/.}
\blfootnote{* All authors contributed equally to this work.}
Code-mixing is a phenomenon which arises mainly in multilingual societies. Multilingual people, who are well versed in their native languages and also English speakers, tend to code-mix using English-based phonetic typing and the insertion of anglicisms in their main language. This linguistic phenomenon poses a great challenge to conventional NLP domains such as Sentiment Analysis, Machine Translation, and Text Summarization, to name a few. In this work, we focus on working out a plausible solution to the domain of Code-Mixed Sentiment Analysis. This work was done as participation in the SemEval-2020 Sentimix Task, where we focused on the sentiment analysis of English-Hindi code-mixed sentences. our username for the submission was "sainik.mahata" and team name was "JUNLP". We used feature extraction algorithms in conjunction with traditional machine learning algorithms such as SVR and Grid Search in an attempt to solve the task. Our approach garnered an f1-score of 66.2\% when tested using metrics prepared by the organizers of the task.
\end{abstract}

\section{Introduction}
\label{intro}
India has a linguistically diverse population due to its long history of foreign acquaintances. English, one of those borrowed languages, became an integral part of the education system and hence gave rise to a population who are very comfortable using bilingualism in their day to day communication. Due to such language diversity and dialects, frequent code-mixing is encountered during conversations. Further, due to the emergence of social media, the practice has become even more widespread. The phenomenon is so common that it is often considered as a different (emerging) variety of the language, e.g., Benglish (Bengali-English) and Hinglish (Hindi-English).

This phenomenon poses a great challenge to the existing domains of Natural Language Processing (NLP) such as Sentiment Analysis as primarily the language technologies, such as parsing, Parts-of-Speech (POS) tagging, etc., are built for English. Furthermore, labeled/annotated data of such category are hard to come by and hence leads to misfiring when using straight-forward machine learning algorithms.

In this work, we participated in SemEval-2020 Sentimix Task\footnote{https://competitions.codalab.org/competitions/20654} and attempted to solve the chore of sentiment analysis of English-Hindi code-mixed sentences.

Initially, our approach includes the use of feature extraction algorithms on the data, procured by the organizers. Thereafter, we used Support Vector Regression coupled with Grid Search algorithm to classify the code-mixed sentences to its respective sentiment class. This approach, when tested using the metrics prepared by the organizers, returned an f1-score of 66.2\%.

The rest of the paper is organized as follows. Section \ref{sec:data} briefly the quantifies the English-Hindi code-mixed data procured by the organizers of the task. Section \ref{sec:methodology} provides a descriptive literature of our proposed approach. This will be followed by the results and concluding remarks in Section \ref{sec:results} and \ref{sec:conclusion}.

\section{Data}
\label{sec:data}
The English-Hindi code-mixed data that was used to train our model was collected from Twitter using the Twitter API, by searching for code-mixed Hindi keywords \cite{patwa2020sentimix}.  The sentiment labels are positive, negative, and neutral. Besides the sentiment labels, the language labels for every word of the code-mixed sentence were also provided. The word-level language tags were ENG (English), HIN (Hindi), and O (Other) for symbols, mentions, and hashtags. 

The organizers provided a trial and a training data set and after adding both, we could gather 17,000 code-mixed instances. We further divided this data into two parts; (i.) 15,000 instances as training data and (ii.) 2,000 instances as validation data.

\section{Methodology}
\label{sec:methodology}
Our approach included converting the given tweets into a sequence of words and then run the Grid Search Cross-Validation algorithm on the processed tweet. Initially, the tweets were pre-processed using methods as done by \cite{garain2019titans} to remove the following:
\begin{enumerate}[noitemsep]
    \item Removing mentions
    \item Removing punctuation
    \item Removing URLs
    \item Contracting white space
    \item Extracting words from hashtags
\end{enumerate}
The last step consisted of taking advantage of the Pascal Casing of hashtags (e.g. \texttt{\#CoronaVirus}). A simple regex can extract all words. This extraction results in better performance mainly because words in hashtags, to some extent, may convey sentiments of hate. They play an important role during the model-training stage.

\subsection{Feature Extraction}
After obtaining clean tweets, various features were extracted by treating them as a sequence of words. Some of the features were manually extracted while some were extracted using pre-existing methodologies like the Bag-of-Words model, GloVe vectors. As our aim is Sentiment Analysis of the texts, so the presence of hate, offense, humor, etc., may have a great influence on the result. The extracted features are listed below. 
\begin{enumerate}[noitemsep]
    \item TF-IDF Vector features: The TF-IDF feature vectors for the texts as a sequence of word vectors.
    \item GloVe Vector features: GloVe vector embeddings for the texts as a sequence of word embeddings.
    \item Humour label and score: Whether a text is humorous or not. If humorous what is its score in the range 0-1.\cite{garain2019humor}
    \item Wordwise sentiment values: List of sentiment values of each word of the text.
    \item Hate and offensiveness labels: Whether the text is offensive or not and if it constitutes hate speech.
    \item Frequency of easy and difficult words: Included as a semantic feature for the texts. \cite{8929231}
\end{enumerate}

\subsection{Learning Model}
Grid search refers to the practice of tuning hyperparameters to determine the most optimal values for a given model. This has a massive significance as the performance of the entire model is highly dependent on the hyperparameter values specified.

The estimator parameter of the Grid Search Cross-Validation process requires the model that has been used for the hyperparameter tuning process. Here the model used is the linear and the RBF kernels of the estimator Support Vector Regression model (SVR). 

This process requires certain parameters to be taken as manual input. The param\_grid parameter itself in turn requires a list of parameters and the range of values for each parameter of the specified estimator. 
The flow diagram has been shown in Fig~\ref{fig:archi}\cite{article}:
\begin{figure}[H]
    \centering
    \includegraphics[scale=0.4]{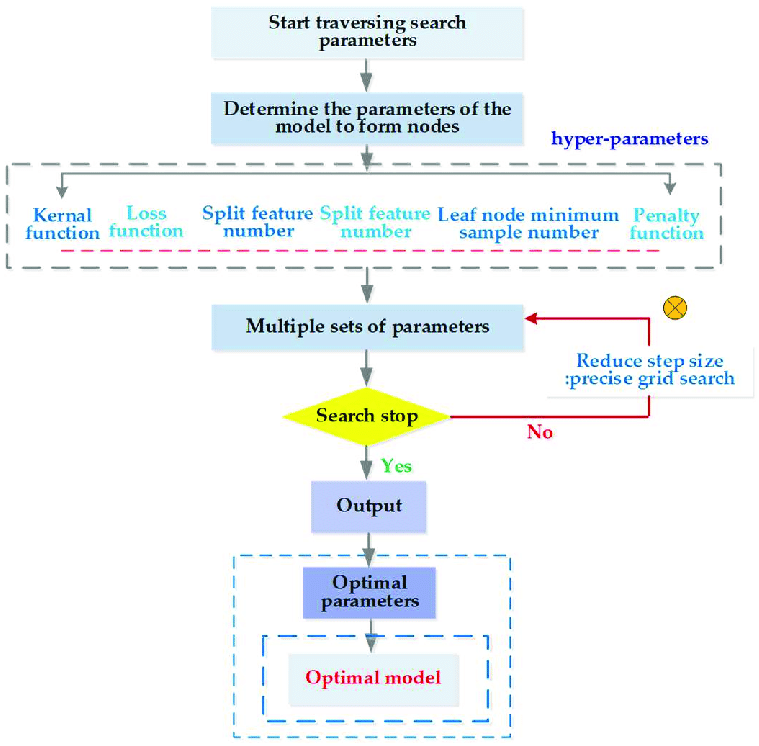}
    \caption{Grid Search parameter optimization overview}
    \label{fig:archi}
\end{figure}
The SVR was fed with parameter values of
\begin{itemize}
    \item C=best\_params["C"] \hspace{5em}$\bullet$ epsilon=best\_params["epsilon"]
    \item gamma=best\_params["gamma"] \hspace{0.5em}$\bullet$ cache\_size=200
    \item coef0=0.1 \hspace{9.3em}$\bullet$ decision\_function\_shape='ovr' 
    \item kernel=['linear','rbf'] \hspace{4.7em}$\bullet$ max\_iter=-1
    \item probability=False \hspace{6.3em}$\bullet$ random\_state=None
    \item shrinking=True \hspace{7.15em}$\bullet$ tol=.001
    \item verbose=False
\end{itemize}

\noindent Class weight and degree were set to Ellipsis.

\noindent The most significant parameters required when working with the RBF kernel of the SVR model were "c", "gamma" and "epsilon". A list of values to choose from has been given to each hyperparameter of the model.

For the GridSearchCV algorithm, parameters like error\_score, iid, param\_grid, pre\_dispatch, refit, return\_train\_score, scoring, and verbose were set to Ellipsis.

A cross validation process is performed in order to determine the hyper parameter value set which provides the best f1-score levels. The parameters for hyper-parameter selection are as follows:
\begin{itemize}
    \item mean\_fit\_time\hspace{3em}$\bullet$ mean\_score\_time
    \item mean\_test\_score\hspace{2.1em}$\bullet$ mean\_train\_score
    \item param\_C\hspace{4.9em}$\bullet$ param\_kernel
    \item params\hspace{5.53em}$\bullet$ rank\_test\_score
    \item split0\_test\_score\hspace{2em}$\bullet$ split0\_train\_score
    \item split1\_test\_score\hspace{2em}$\bullet$ split1\_train\_score
    \item split2\_test\_score\hspace{2em}$\bullet$ split\_train\_score
    \item std\_fit\_time\hspace{3.93em}$\bullet$ std\_score\_time
    \item std\_test\_score\hspace{3em}$\bullet$ std\_train\_score
\end{itemize}
 Experimentation has been performed thoroughly and the parameters giving the best results have been accepted.\\

\section{Results}
\label{sec:results}
The metric for evaluating the participating systems was as follows. The organizers used F1 averaged across the positives, negatives, and the neutral. The final ranking was based on the average F1 score. Our submitted system garnered an F1 score of 66.2\%.
 The detailed results are shown in Table \ref{tab1:results-A-open}:
\begin{table}[H]
\center
\begin{tabular}{|l|l|l|l|l|}
\hline
\textbf{Class}    & \textbf{Precision} & \textbf{Recall} & \textbf{F1-score} & \textbf{Support} \\ \hline
\textbf{negative} & 0.68               & 0.68            & 0.68              & 900              \\ \hline
\textbf{neutral}  & 0.57               & 0.59            & 0.58              & 1100              \\ \hline
\textbf{positive} & 0.75               & 0.72            & 0.74              & 1000               \\ \hline
\textbf{Macro avg.}      & \textbf{0.66}      & \textbf{0.66}   & \textbf{0.662}     & \textbf{3000}     \\ \hline
\end{tabular}
 \caption{Class wise full result metrics}
\label{tab1:results-A-open}
 \end{table}



\section{Conclusion}
\label{sec:conclusion}
In the current work, we attempted to solve the problem of Sentiment Analysis of code-mixed English-Hindi data, while participating in the SemEval shared task. Our system was based on using traditional machine learning algorithms coupled with Beam Search Cross-Validation. Our system, when evaluated by the organizers garnered an F1 score of 0.662. There was an option of developing an unconstrained system, but we only used the provided data to develop the system. As future work, we would like to increase this data, use state-of-the-art Neural Network architectures on this data, taking into advantage the concept, matrix and embedded language, SentiWordNet, and other NLP features.
\nocite{*}
\bibliography{coling2020}
\bibliographystyle{coling}

\end{document}